\documentclass[letterpaper]{article} 
\usepackage{aaai2026}  
\nocopyright 
\usepackage{tcolorbox}
\usepackage{xcolor}
\usepackage{times}  
\usepackage{helvet}  
\usepackage{courier}  
\usepackage[hyphens]{url}  
\usepackage{graphicx} 
\usepackage{natbib}  
\usepackage{caption} 
\usepackage{booktabs}
\usepackage{threeparttable}

\tcbuselibrary{minted, breakable, skins, listings}

\usepackage{algorithm}

\usepackage{algorithmic}

\usepackage{newfloat}
\usepackage{listings}
\DeclareCaptionStyle{ruled}{labelfont=normalfont,labelsep=colon,strut=off} 
\floatstyle{ruled}
\newfloat{listing}{tb}{lst}{}
\floatname{listing}{Listing}
\title{One Agent to Serve All: a Lite-Adaptive Stylized AI Assistant for Millions of Multi-Style Official Accounts}
\author {
    Xingyu Fan, Feifei Li, Wenhui Que\textsuperscript{*}, Hailong Li
}
\affiliations {
    WeChat, Tencent Inc., Beijing, China\\
    \texttt{\{fanxfan, niyali, victorque, aloneli\}@tencent.com}
}

\newcommand{\tool}{WeStar}

\usepackage{bibentry}

\begin{document}

\maketitle

\begin{abstract}
Conversational agents deployed in industrial-scale official account platforms must generate responses that are both contextually grounded and stylistically aligned—requirements that existing methods struggle to meet. Chain-of-thought (CoT) prompting induces significant latency due to multi-turn reasoning; per-account fine-tuning is computationally prohibitive; and long prompt-based methods degrade the model’s ability to grasp injected context and style.
In this paper, we propose \tool{}, a lite-adaptive framework for stylized contextual question answering that scales to millions of official accounts. \tool{} combines context-grounded generation via RAG with style-aware generation using Parametric RAG (PRAG), where LoRA modules are dynamically activated per style cluster.
Our contributions are fourfold: (1) We introduce \tool{}, a unified framework capable of serving large volumes of official accounts with minimal overhead. (2) We propose a multi-dimensional, cluster-based parameter sharing scheme that enables compact style representation while preserving stylistic diversity. (3) We develop a style-enhanced Direct Preference Optimization (SeDPO) method to optimize each style cluster’s parameters for improved generation quality. (4) Experiments on a large-scale industrial dataset validate the effectiveness and efficiency of \tool{}, underscoring its pracitical value in real-world deployment.
\end{abstract}
\section{Introduction}
Inspired by the outstanding capabilities of large language models in question-answering tasks, the industry has adopted conversational agents that dialogue with users in many tasks, such as AI in games ~\cite{gameai}, voice assistants ~\cite{voice}, and official account assistants.
Our Industrial Official Accounts serve as a powerful communication channel for individuals, media outlets, enterprises, government bodies, and other organizations, enabling them to disseminate information in article form within our ecosystem. 
Users can interact with these articles by leaving comments, which are often replied to directly by the author. Additionally, users may pose questions to the Official Account’s intelligent assistant via the chat interface, expecting responses that are both context-grounded and style-aware which is grounded in the author’s published articles and reflective of the author’s personal communication style.
While the articles provide rich, factual content, they are typically formal and do not capture the author’s conversational tone. In contrast, the author’s replies to user comments offer a more authentic and fine-grained reflection of their stylistic preferences in interactive settings. 
Motivated by this observation, we treat articles as the source of question-specific knowledge and leverage the author's historical comment replies as the basis for style-specific knowledge to address the \textbf{Stylized Contextual Question Answering} problem.

To tackle the problem, existing approaches can be broadly categorized into three paradigms: \textbf{fine-tuning-based }methods, which directly adapt the model on style-specific data; \textbf{CoT-based }methods, which employ multi-step prompting to decompose and solve the task; and \textbf{prompt-based }methods, which inject both knowledge-specific and style-specific knowledge into a single prompt. However, each of these comes with limitations in scalability, efficiency, or effectiveness when deployed at industrial scale.

In recent years, fine-tuning-based approaches have shown strong performance in the domain of stylized text generation~\cite{sftmethod1}. A widely adopted strategy involves supervised fine-tuning (SFT) on customized style-specific corpora, enabling large language models (LLMs) to adjust their output distributions by updating model parameters accordingly.
However, fine-tuning remains a significant bottleneck in many real-world applications. For instance, 
in our public account assistant tasks, a distinct model must be fine-tuned and maintained for each individual author to ensure stylistic consistency. This process incurs substantial time and computational costs, severely limiting scalability.

Chain-of-thought (CoT)-based methods offer a promising direction for mitigating the computational and deployment burden associated with stylized text generation. A straightforward solution is to decompose the stylized contextual question answering task into two sequential sub-tasks: (1) generating a contextually relevant answer, followed by (2) applying a text style transfer model to adapt the response to the target style.
While conceptually simple, this two-stage pipeline introduces practical limitations. Specifically, invoking a large language model (LLM) twice not only increases computational overhead but also introduces latency, which can significantly degrade the user experience. 

Therefore, prompt-based methods presents a more practical solution to the stylized contextual question answering task. By incorporating retrieved question-specific articles and customized style-specific corpora into the input of an end-to-end LLM, the system injects external stylistic and semantic knowledge directly into the model's context window, enabling dynamic adaptation without parameter updates.
However, injecting knowledge from multiple sources via input prompts inevitably leads to increased context length. This extended input not only introduces additional computational overhead and latency during inference, but also degrades the LLM’s ability to effectively understand and utilize the injected information—particularly in scenarios requiring complex reasoning~\cite{promptlimit}.

To address these challenges, we propose \textbf{\tool{}}, a novel framework to build one lite-adaptive style-ware and context-grounded agent capable of serving millions of multi-style official accounts. Before online inference, \tool{} first performs fine-grained style labeling over each public account author's corpus across multiple stylistic dimensions. Authors with similar style are then grouped into clusters, and each cluster is associated with a shared set of stylized model parameters  trained via style-enhanced Direct Preference Optimization (SeDPO). This cluster-based parameter sharing enables compact storage of stylistic knowledge and supports scalable deployment across millions of public account authors.
During inference, \tool{} incorporates question-specific knowledge (i.e., retrieved articles) into the input prompt to enrich the model’s domain understanding and enhance its question-answering capabilities. In parallel, instead of relying exclusively on prompt-based knowledge injection, \tool{} adopts a parameter injection approach inspired by PRAG ~\cite{prag}, where style-specific knowledge is embedded directly into the model parameters. This dual-channel design not only enhances stylistic consistency but also substantially reduces prompt length, thereby mitigating context overflow and improving inference efficiency.

To evaluate the effectiveness of \tool{} in real-world industrial scenarios, we conducted experiments on a large-scale public account dataset. \tool{} achieves state-of-the-art performance across four customized evaluation dimensions, including contextual alignment, question relevance, stylistic strength and fluency. These results demonstrate the practical applicability of our framework in stylized contextual generation tasks.

Our main contributions are summarized as follows:
\begin{itemize}
    \item We propose \tool{}, a novel framework to build one lite-adaptive stylized contextual question-answering agent capable of serving millions of official accounts.
    \item We introduce a multi-dimensional style-specific cluster-based parameter sharing approach that enables compact parameter storage while preserving a broad spectrum of stylistic knowledge, facilitating scalable deployment across diverse official account authors.
    \item We leverage a style-enhanced Direct Preference Optimization (SeDPO) strategy to train the parameter representations for each style cluster, thereby improving the model’s capacity for style-aware generation.
    \item We conduct evaluations on a large-scale industrial dataset and validate the proposed method across four key metrics, demonstrating its practical value and profitability in real-world applications.
\end{itemize}

\section{Related Work }
\subsection{Text Style Transfer}
Text style transfer (TST) is a different but related task compared to our work, which aims to alter the stylistic attributes of a given text while preserving its original content. 
Early research focused on rule-based and statistical methods. With the rise of deep learning, neural methods became dominant, particularly unsupervised frameworks leveraging content-style disentanglement or reinforcement learning~\cite{tst2, tst1}. 
More recent work investigates the potential of contrastive learning and pattern mining~\cite{tst3}.
LLMs also bring new paradigms to TST. Reif et al.\cite{tst4} and Mukherjee et al.\cite{tst5} explored zero-shot and few-shot style transfer via prompt engineering, revealing LLMs’ surprising generalization ability. Ostheimer et al.~\cite{tst6} further investigated LLM-based evaluation, demonstrating strong correlation with human judgments. 
\subsection{Stylized Answer Generation}
Several studies tackle stylized answer generation using fine-tuned large language models or style-controlled decoding strategies~\cite{scqa1, scqa2}. 
To better ground stylized generation in external knowledge, Sun et al.\cite{scqa3} introduce disentangled template rewriting in knowledge-grounded scenarios. 
Additionally, Feature-Guided Knowledge Augmentation\cite{scqa4} retrieves stylistic sentences to guide content planning and uses contrastive learning to enhance fluency and style control. 
Despite these advances, most prior work operates on relatively small-scale or synthetic style corpora, which limits their applicability in real-world, large-scale scenarios. Recent efforts have begun to address scalability ~\cite{scqa5, scqa6}, but none match the scale and complexity of the deployment setting considered in our work.

\section{Method}
Given a user-issued question $Q$, a retrieved context-specific knowledge $C$ relevant to $Q$, and a retrieved style-specific knowledge $S$ representing a target response style, the objective is to generate a style-aware and context-grounded answer $A$ such that
\begin{itemize}
    \item $A$ correctly addresses the information need expressed in $Q$, grounded in the content of $C$, and
    \item $A$ adheres to the stylistic characteristics exemplified by $S$.
\end{itemize}
To address the challenges of stylized contextual question answering, \tool{} adopts a dual-injection approach: it injects question-specific knowledge into the prompt and style-specific knowledge into the model parameters. In this section, we first describe how \tool{} clusters authors with similar style and trains shared stylize-specific parameters for each style cluster. We then present the online inference mechanism of \tool{}, which dynamically composes responses by combining the retrieved content and the appropriate style-specific parameters. 

\begin{figure*}[t]
  \centering
  \includegraphics[width=18cm]{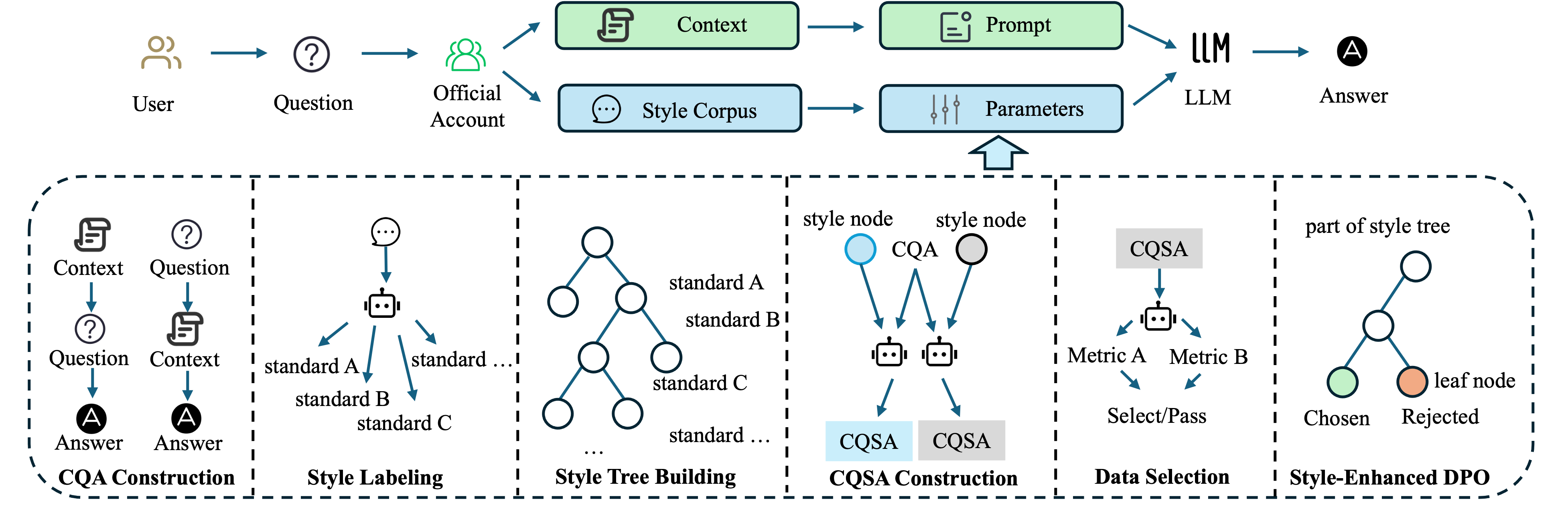}
  \caption{Overview of \tool{}. }
  \label{fig:Overview}
\end{figure*}


\subsection{CQA Construction}

To construct high-quality CQA (\textit{Context, Question, Answer}) triplets for training, we adopt two complementary strategies: a forward-thinking method that prompts a large language model $M$ to generate questions and answers based on given article segments, and a bottom-up method that prompts $M$ to simulate realistic user roles and queries based on the domain of each public account, followed by context retrieval and answer generation with $M$. The forward-thinking approach offers high scalability, while the bottom-up strategy introduces domain-relevant and user-intent-aligned queries, ensuring a better match to real-world QA workflows. Together, these methods yield a CQA dataset that balances diversity, difficulty, and contextual grounding. Full prompt templates and examples are provided in the \textbf{Appendix}.

\subsection{Style Labeling}
We design twelve style classification standards, organized across four stylistic dimensions:
\begin{itemize}
\item \textbf{Semantic level:} intention type, degree of authority
\item \textbf{Grammatical level:} omitted features, use of inversion, use of passive voice
\item \textbf{Syntactic level:} sentence complexity, rhetorical features, cohesion mechanisms
\item \textbf{Lexical level:} lexical complexity, emotional polarity, frequency of emojis, and degree of formality
\end{itemize}

For each official account author, we perform fine-grained labeling of their style corpus using LLM $M$, based on the twelve predefined classification standards. Specifically, for every QA pair in the corpus, $M$ generates candidate labels for each stylistic dimension. We then aggregate these annotations by taking the majority label for each standard across all QA pairs within the author's corpus. These aggregated labels serve as the author's stylistic profile and provide the foundation for subsequent style tree building. The detailed prompt templates are provided in the \textbf{Appendix}.

\subsection{Style Tree Building}
\label{sec:treebuilding}
\begin{algorithm}[t] 
\caption{style tree building}
\label{code:buildtree}
\textbf{Input:} $\mathcal{C}$ – a set of style corpora \\
\textbf{Input:} $\mathcal{S}$ – a set of style classification standards \\
\textbf{Output:} $\mathcal{T}$ – a hierarchical style tree

\vspace{1mm}
\textbf{1}  \hspace{2mm}     Initialize tree $\mathcal{T}$ with a root node containing \\
\hspace*{5mm}         all corpora $\mathcal{C}$ \\
\textbf{2}  \hspace{2mm}     Initialize queue $Q \leftarrow$ \{root node\} \\
\textbf{3}  \hspace{2mm}     Set corpus size threshold $k$ \\
\textbf{4}  \hspace{2mm}    \textbf{For each} style standard $s \in \mathcal{S}$: \\
\textbf{5}  \hspace{6mm}     \textbf{While} $Q$ is not empty: \\
\textbf{6}  \hspace{10mm} $n$ $\leftarrow$ $Q.front()$ \\
\textbf{7}  \hspace{10mm} $Q.pop()$ \\
\textbf{8}  \hspace{10mm} \textbf{If} node $n$ can be split into child nodes $\{n_1, ..., n_m\}$ 
\hspace*{13mm} by $s$ \textbf{and} each child $n_i$ satisfies $|\mathcal{C}_{n_i}| > k$, \textbf{then}:\\
\textbf{9}  \hspace{14mm} Expand node $n$ in $\mathcal{T}$ into children $\{n_1, ..., n_m\}$ \\
\textbf{10} \hspace{4.2mm} Push all new leaf nodes in $\mathcal{T}$ into queue $Q$ \\
\textbf{11} \hspace{0.2mm} Return $\mathcal{T}$
\end{algorithm}
    
Following the style labeling stage, we construct a hierarchical style clustering tree to group the authors with similar style into the same cluster. 
The construction process, outlined in Algorithm~\ref{code:buildtree}, proceeds by traversing the predefined set of style labeling standards $\mathcal{S}$ in a specified hierarchical order. For each leaf node in the current tree, if the associated style corpus can be partitioned by a label standard $s \in \mathcal{S}$—and each resulting subset contains more than $k$ examples—then the node is expanded into multiple child nodes, each corresponding to a distinct stylistic subgroup defined by $s$. Upon completion of the algorithm, each leaf node in the resulting style clustering tree represents a stylized cluster of the original corpus $\mathcal{C}$. The path from a given leaf node to the root captures its cumulative stylistic characteristics.

In this way, authors with similar stylistic characteristics are grouped into the same clusters. This hierarchical organization facilitates style-specific parameter training by enabling parameter sharing among authors within the same cluster, thus reducing both training costs and storage overhead. Moreover, for authors with limited stylistic data, this clustering mechanism allows them to benefit from the shared parameters of stylistically similar authors, thereby enhancing generalization and mitigating the data scarcity issue.

\subsection{CQSA Construction}
Many recent studies have demonstrated the strong capabilities of LLMs in text style transfer(TST) ~\cite{tst5}. Building on this foundation, we leverage LLMs to transform standard CQA instances into CQSA \textit{(Context, Question, Stylized Answer)} instances that align with the target style for each cluster.

For each corpus within the style cluster derived from the style tree, we prompt LLM $M$ to rewrite the original answer—while preserving the factual correctness of the response—into a form that conforms to the style of the target corpus. The prompt is composed of the following elements: 
\begin{enumerate}
    \item the input context and question from the original CQA pair,
    \item the original answer to be rewritten,
    \item the full set of twelve style classification standards and labels associated with the target cluster,
    \item $m$ in-context examples selected from the same cluster, each consisting of a user comment and the corresponding author reply.
\end{enumerate} 
To ensure stylistic consistency, the in-context examples are sampled uniformly at random from each author in the same cluster. The detailed prompts are presented in the \textbf{Appendix}.
\subsection{Data Selection}
Inspired by metric-based RLHF ~\cite{mdpo}, we apply metric-based constraints during data construction to better align the model's outputs with the target stylistic expectations, while simultaneously reducing the likelihood of hallucinations.
Following previous works~\cite{mdpo, tst6}, we conduct automated evaluation on each CQSA instance using LLM $M$ across four key dimensions: Contextual Alignment (\textbf{C–A}), Question Relevance (\textbf{Q–A}), Stylistic Strength (\textbf{S–A}) and \textbf{Fluency}, which is presented in the \textbf{Appendix}. We will further discuss these metrics in Section \textbf{Metrics}. We aggregate the scores across these four metrics and select the top 10,000 CQSA instances as high-quality samples for subsequent style-specific parameter training.
\subsection{Style-Enhanced DPO}
Prior to this stage, we have obtained high-quality CQSA instances for each style cluster to support style-specific parameter training. In this step, we adopt style-enhanced DPO (SeDPO) to train the parameters. Specifically, for a given style cluster, the top 10,000 CQSA instances are used as chosen samples. We construct the corresponding rejected sample for each chosen sample by retrieving answers to the same question with high stylistic similarity—such as sibling nodes in the style tree—while ensuring that these answers differ in a certain style label.
This construction aligns with the principle of controlled variable experimentation: when the negative sample shares most contextual and semantic features with the positive sample, the model is encouraged to focus more on the fine-grained stylistic distinctions. Consequently, this training setup facilitates more effective learning of style-specific behaviors within each stylistic cluster.

We adopt the same parameterization and injection paradigm as PRAG~\cite{prag}, employing LoRA~\cite{lora} as our fine-tuning and parameter storage strategy. This design enables each style cluster to be associated with an independently trained set of low-rank adaptation parameters, which allows the model to encode style-specific behaviors in a parameter-efficient manner, enabling scalable and flexible deployment across a wide range of stylistic clusters without the need to train or deploy the full base model.

\subsection{Online Inference} 
After training a set of style-specific parameters for each style cluster, \tool{} applies them during online inference to deliver style-aware, context-grounded responses at scale.

\tool{} performs inference by jointly injecting question-specific and style-specific knowledge into the generation process. Specifically, question-specific article segments are inserted into the input prompt, providing contextual grounding. Simultaneously, style-specific LoRA parameters corresponding to the author’s style cluster are retrieved via a PRAG (Parametric Retrieval-Augmented Generation) manner and injected into the model’s parameter space. This dual-injection strategy enables \tool{} to generate responses that are both contextually relevant and stylistically aligned, while maintaining scalability across a large number of public account authors.
\section{Experiments}
\subsection{Experimental Settings}
\subsubsection{Dataset}

To the best of our knowledge, there exists no public dataset that simultaneously provides articles, user queries, and large-scale stylized replies from millions of authors. Therefore, we directly evaluate our method using proprietary data from a widely-used, real-world industrial official accounts platform. We deployed the \tool{} framework in this environment and conducted evaluations across ten representative style clusters, constructed using the methodology detailed in Section~\textbf{Style Tree Building}. Each cluster contains real user comments and the corresponding author replies collected before July 2025, serving as style reference for stylized question answering. The contextual retrieval corpus consists of all historical articles published by the authors prior to the same date.  To perform evaluation, we constructed a test set of 2,000 instances using the same methodology described in Section \textbf{CQA Construction} and Section \textbf{CQSA Construction}. To better simulate real-world deployment scenarios, we further supplemented the test set with 3,000 user-generated information-seeking questions collected from live interactions. In total, the evaluation dataset comprises 5,000 queries, covering both controlled and in-the-wild user intents.

\subsubsection{Metrics}
To ensure consistency and comparability, we employ the same four evaluation metrics as described in Section~\textbf{Data Selection} to assess model performance. These metrics have been widely adopted in previous studies~\cite{tst6, mdpo}:
\begin{itemize}
\item \textbf{Contextual Alignment (C–A)}: the degree of semantic consistency between the generated answer and the retrieved context;
\item \textbf{Question Relevance (Q–A)}: the extent to which the answer accurately addresses the core intent of the input question;
\item \textbf{Stylistic Strength (S–A)}: the degree to which the answer adheres to the target stylistic attributes;
\item \textbf{Fluency}: the grammaticality and naturalness of the generated response.
\end{itemize}

To obtain consistent and scalable evaluations, we employed DeepSeek-R1 ~\cite{r1} to rate each output along the four dimensions, using standardized evaluation prompts. The exact prompt formulations are provided in the \textbf{Appendix}.
\subsubsection{Baselines}
As discussed in the \textbf{Introduction}, stylized contextual question answering in industrial settings—such as official account platforms—requires a fully end-to-end LLM pipeline. This constraint renders multi-step or sequential prompting approaches infeasible due to latency and system complexity. To the best of our knowledge, there exists no prior method capable of supporting scalable stylized contextual question answering across millions of official account authors with distinct stylistic preferences.

To evaluate the effectiveness of \tool{} under this challenging setting, we compare it against five baseline methods, covering diverse methodological paradigms: (1) two prompting-based approaches, (2) two SFT variants from \tool{}, and (3) one variant of DPO:

\begin{itemize}
    \item \textbf{R1-Prompt}. We adopt DeepSeek-R1~\cite{r1}, a recently released open-source large language model that has demonstrated strong performance across a wide range of reasoning, instruction-following, and language understanding benchmarks, making it one of the most representative open-source LLMs available to date. In this prompt-based baseline, we use DeepSeek-R1 as the base model and construct the input prompt by injecting four key elements: (1) the user’s question, (2) a set of retrieved articles from the corresponding official account, (3) recent high-quality author replies that reflect the account’s stylistic preferences, and (4) a system-level instruction aligned with the intelligent assistant scenario. The model then generates an answer conditioned on both contextual knowledge and stylistic cues. We consider R1-Promp to be a strong representative of the \textbf{prompt-based paradigm}, as it leverages a state-of-the-art base model and simulates realistic deployment constraints using only prompt engineering without any task-specific fine-tuning.
    
    \item \textbf{SFT-Prompt}. While DeepSeek-R1 exhibits strong performance, its large parameter size incurs substantial inference latency, making it less practical for real-time applications. To address this issue, we construct a second baseline, SFT-Prompt, based on the Qwen3-32B model~\cite{qwen3}, which offers a better trade-off between performance and latency in industrial deployment scenarios. Specifically, we first fine-tune Qwen3-32B using 10,000 CQA instances to enhance its ability to perform contextual reasoning ~\cite{nv}. After fine-tuning, we adopt the same prompt injection strategy as in R1-Prompt, incorporating the user query, retrieved articles, representative stylistic replies, and system-level task instructions. This setting represents the \textbf{SFT-then-prompting paradigm} under similar model scale constraints and serves as a strong baseline for evaluating methods that enhance LLMs via lightweight supervised adaptation followed by prompting.

    \item \textbf{LoRA-SFT}. This baseline adopts the same online-inference paradigm as proposed in \tool{}. For each style corpus, we apply supervised fine-tuning using LoRA to train a set of style-specific parameters. At inference time, these parameters are dynamically injected into the base model following the PRAG parameter loading mechanism. The training set consists of 10,000 randomly selected CQSA instances sampled prior to the data selection step described in Section~\textbf{Data Selection}.

    \item \textbf{LoRA-SFT-S}. This variant builds on LoRA-SFT, with the only difference being the training data quality. Instead of random sampling, we train the stylized parameters using the top 10,000 high-quality CQSA instances selected through the metric-based data selection process introduced in Section~\textbf{Data Selection}. This setting enables us to assess the impact of curated, high-quality training data on the effectiveness of style-specific parameter learning.

    \item \textbf{\tool{}$_{MDPO}$}. This baseline shares the same online inference setup and training paradigm as \tool{}, but differs in the construction of rejected samples used for DPO. Instead of using style-aware, high-quality CQSA data selected via the procedures outlined in Sections~\textbf{CQSA Construction} and~\textbf{Data Selection}, the rejected responses are distilled from the base LLM by prompting it with corresponding \textit{(C, Q)} pairs, without further quality filtering. To enhance learning signals, we apply a metric-guided DPO strategy inspired by MDPO~\cite{mdpo}, where we select rejected samples that exhibit the greatest deviation from the chosen instances across the four evaluation metrics.
\end{itemize}
\subsubsection{Implementation Details}
We adopt Qwen3-32B as the base language model. Additionally DeepSeek-R1 is chosen as the the auxiliary LLM $M$ throughout our framework. During the \textbf{CQA Construction} phase, we prompt $M$ to generate \textbf{3} representative user roles for each official account domain, and subsequently generate \textbf{3} domain-relevant questions per role to construct a diverse question set.
In the \textbf{Style Tree Building} stage, we set the corpus size threshold \textbf{k=100} to ensure sufficient stylistic representation in each node before further partitioning.
For \textbf{Style-enhanced DPO}, we employ LoRA with a rank of 16 and train for one epoch, enabling efficient parameter adaptation for each style corpus while maintaining training scalability.

\subsection{Main Results}

\subsubsection{\tool{} vs. Prompt-Based Methods}
\begin{figure}[t]
    \centering
    \includegraphics[width=\linewidth]{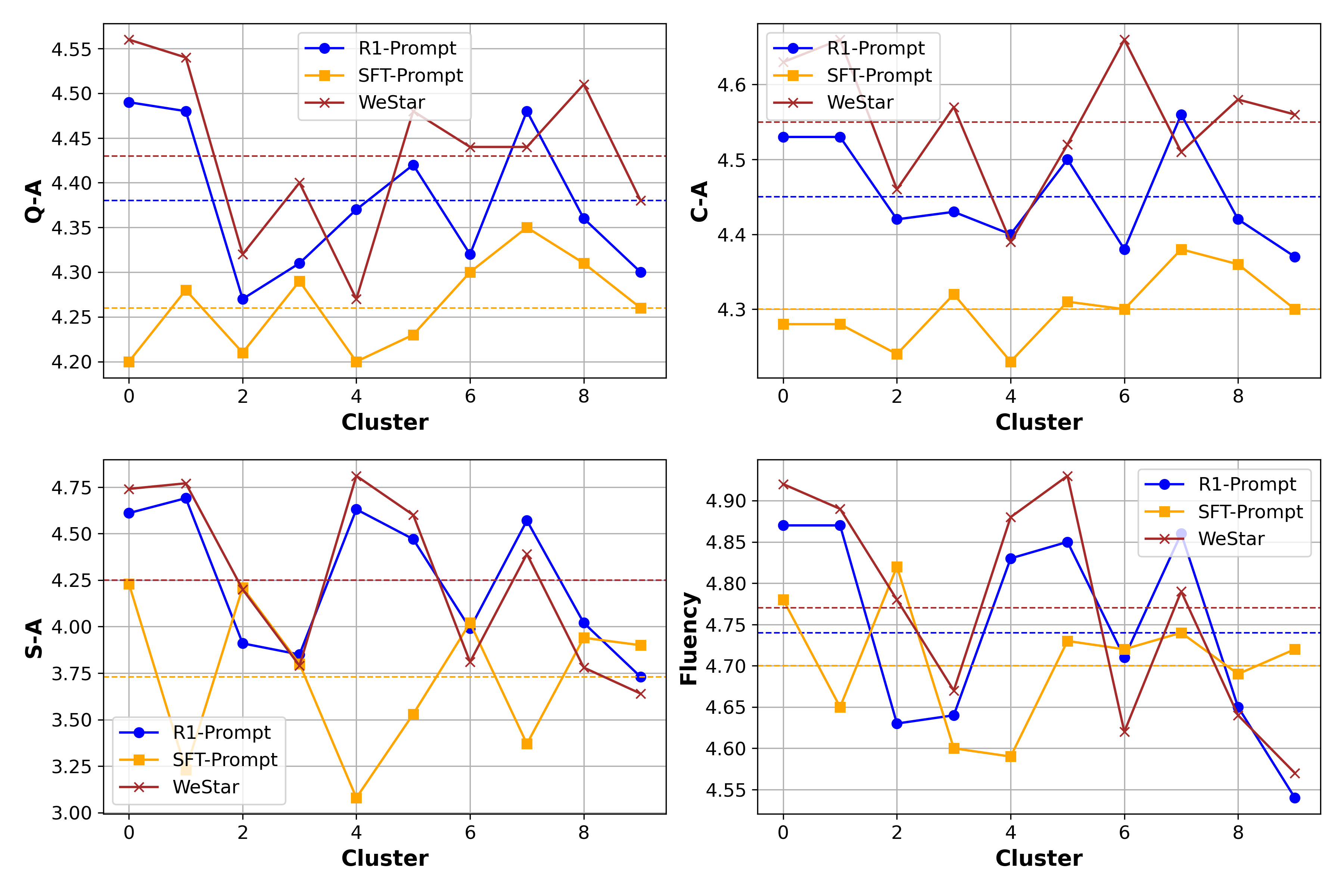}
    \vspace{-4mm}
    \caption{Results of \tool{} vs. Prompt-Based Methods}
    \label{fig:ourvsprompt}
    \vspace{-4mm}
\end{figure}

Figure~\ref{fig:ourvsprompt} illustrates the comparative performance between \tool{} and prompt-based methods across all four evaluation metrics mentioned above. Specifically, the top-left, top-right, bottom-left, and bottom-right subfigures respectively show the per-cluster comparison results between \tool{}, R1-Prompt and SFT-Prompt along each metric. The x-axis represents the identifiers of different style corpus clusters, while the y-axis corresponds to the metric scores achieved by each method on those clusters.
More detailed numerical results are presented in Table~\ref{tab:main_result}. The first column lists the style corpus clusters and the associated evaluation metrics. Columns 2, 3, and 7 report the performance of R1-Prompt, SFT-Prompt, and \tool{}, respectively.

From the results, we observe that \tool{} consistently outperforms both prompting-based approaches on Q–A, C–A , and Fluency metrics on average. This improvement stems from the limitations of prompt-based methods when dealing with long input sequences. These approaches inject both the retrieved articles and the full style corpus into the prompt, leading to significant context length expansion, which in turn challenges the LLM’s attention window. As a result, the model’s ability to accurately comprehend and utilize the injected information is weakened.
For the Stylistic Strength (S–A) metric, only R1-Prompt achieves performance comparable to \tool{} on average. We attribute this to two main reasons:
(1) Prompt-based methods can flexibly incorporate the original author’s entire style corpus directly into the prompt, whereas \tool{} utilizes corpora from style-similar authors identified via the style tree;
(2) The base model used in R1-Prompt, DeepSeek-R1, contains significantly more parameters than our base model, Qwen3-32B, which may contribute to its stronger stylistic generation performance. Consequently, \tool{} matches but does not surpass R1-Prompt on the S–A dimension.
\begin{table*}
\setlength{\tabcolsep}{5.5pt}
\renewcommand{\arraystretch}{0.8}
\centering
\caption{Main Results}
\label{tab:main_result}
\begin{tabular}{l|cc|cc|cc}
\multicolumn{1}{c|}{} & \multicolumn{2}{c|}{\textbf{Prompt-Based Methods}} & \multicolumn{2}{c|}{\textbf{SFT-Based Methods}} & \multicolumn{2}{c}{\textbf{DPO-Based Methods}}\\
\toprule
Dataset \& Metrics & R1-Prompt & SFT-Prompt & LoRA-SFT & LoRA-SFT-S & \tool{}$_{MDPO}$ & \tool{} \\
\midrule
cluster 0 Q-A & 4.49 & 4.20 & 4.34 & 4.44 & \underline{4.50} & \textbf{4.56} \\
cluster 0 C-A & 4.53 & 4.28 & 4.37 & 4.51 & \underline{4.62} & \textbf{4.63} \\
cluster 0 S-A & 4.61 & 4.23 & 4.40 & \underline{4.73} & 4.69 & \textbf{4.74} \\
cluster 0 Fluency & 4.87 & 4.78 & 4.91 & 4.91 & \textbf{4.96} & \underline{4.92} \\
cluster 1 Q-A & 4.48 & 4.28 & 4.42 & 4.44 & \textbf{4.58} & \underline{4.54} \\
cluster 1 C-A & 4.53 & 4.28 & 4.48 & 4.50 & \textbf{4.70} & \underline{4.66} \\
cluster 1 S-A & 4.69 & 3.23 & 4.68 & \textbf{4.78} & 4.76 & \underline{4.77} \\
cluster 1 Fluency & 4.87 & 4.65 & \textbf{4.92} & \underline{4.90} & \underline{4.90} & 4.89 \\
cluster 2 Q-A & 4.27 & 4.21 & 4.15 & \underline{4.33} & \textbf{4.37} & 4.32 \\
cluster 2 C-A & 4.42 & 4.24 & 4.29 & 4.41 & \textbf{4.48} & \underline{4.46} \\
cluster 2 S-A & 3.91 & \textbf{4.21} & 4.02 & 4.05 & 4.11 & \underline{4.20} \\
cluster 2 Fluency & 4.63 & \textbf{4.82} & 4.77 & 4.71 & 4.75 & \underline{4.78} \\
cluster 3 Q-A & 4.31 & 4.29 & \textbf{4.42} & 4.38 & \underline{4.40} & \underline{4.40} \\
cluster 3 C-A & 4.43 & 4.32 & \underline{4.51} & 4.46 & 4.49 & \textbf{4.57} \\
cluster 3 S-A & \textbf{3.85} & \underline{3.80} & 3.62 & 3.75 & 3.72 & 3.79 \\
cluster 3 Fluency & \underline{4.64} & 4.60 & 4.58 & 4.59 & 4.56 & \textbf{4.67} \\
\midrule
\multicolumn{7}{c}{Due to the space limit, we put the detailed results of other six clusters in the \textbf{Appendix}.} \\ 
\midrule
average Q-A & 4.38 & 4.26 & 4.35 & 4.41 & \textbf{4.44} & \underline{4.43} \\
average C-A & 4.45 & 4.30 & 4.43 & 4.49 & \underline{4.52} & \textbf{4.55} \\
average S-A & \textbf{4.25} & 3.73 & 3.92 & \underline{4.22} & 4.20 & \textbf{4.25} \\
average Fluency & 4.75 & 4.70 & 4.73 & \textbf{4.77} & \underline{4.76} & \textbf{4.77} \\
\bottomrule
\end{tabular}
\begin{tablenotes}
\small
\item \textbf{Bold} indicates the best result; \underline{Underline} indicates the second-best.
\end{tablenotes}
\end{table*}

\begin{figure}[t]
    \centering
    \includegraphics[width=\linewidth]{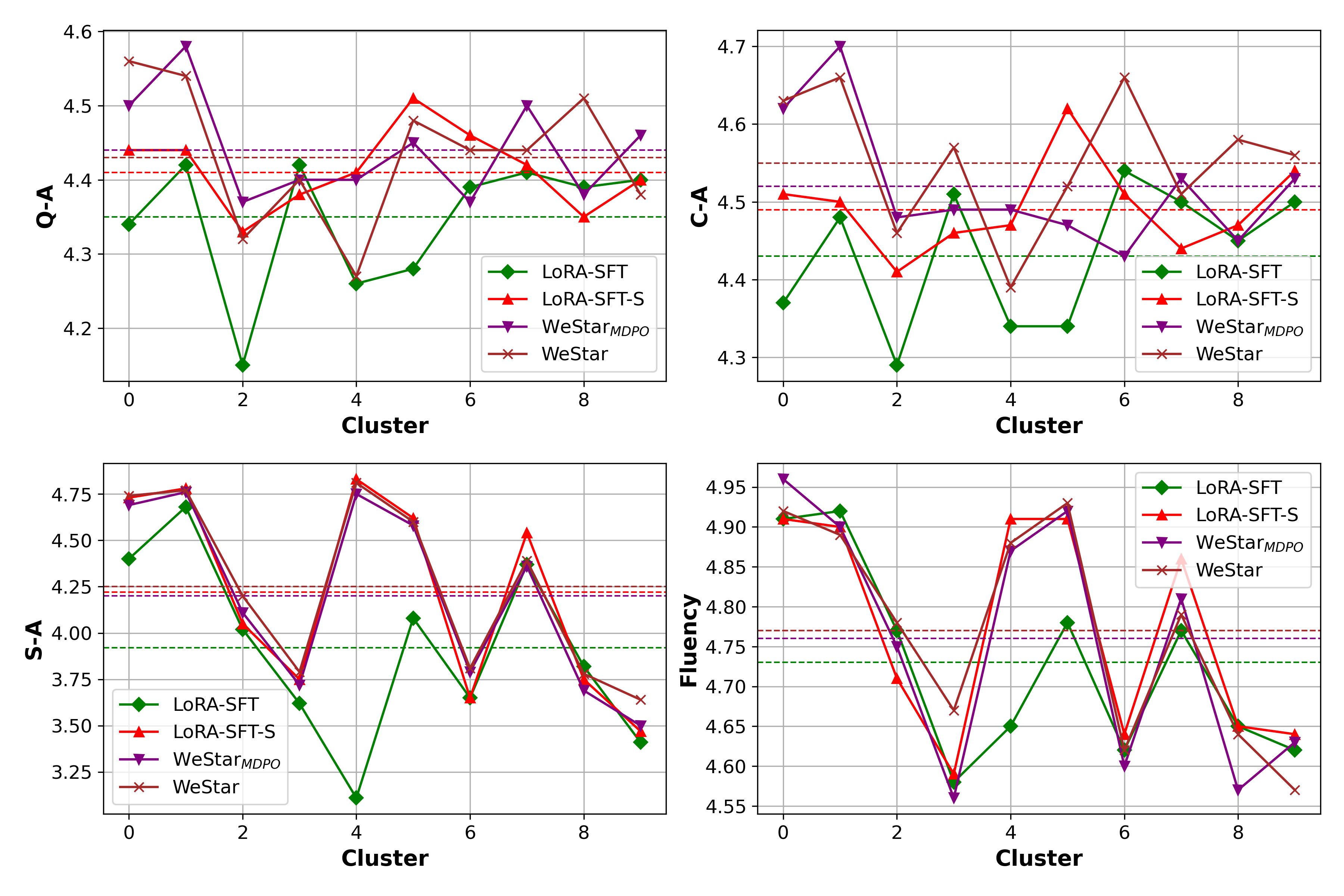}
    \vspace{-4mm}

    \caption{Results of \tool{} vs. Variants of \tool{}}
    \label{fig:variant}
    \vspace{-4mm}
\end{figure}

\subsubsection{\tool{} vs. Variants of \tool{}}
Figure~\ref{fig:variant} presents a comparative analysis of \tool{} and its three training variants: LoRA-SFT, LoRA-SFT-S, and \tool{}$_{MDPO}$—across all four evaluation metrics. The detailed results can be found in Table~\ref{tab:main_result}, where the 4th to 7th columns correspond to the respective performances of LoRA-SFT, LoRA-SFT-S, \tool{}$_{MDPO}$, and \tool{}.

Among these, LoRA-SFT underperforms the other three methods across all metrics on average. This is primarily due to the lack of metric-based data filtering in its training phase. In contrast, the remaining three approaches benefit from the metric-based data selection pipeline, which ensures higher-quality CQSA instances and leads to more accurate style-aware and context-grounded responses.

Overall, \tool{} achieves the best average performance, outperforming both LoRA-SFT-S and \tool{}$_{MDPO}$, except on the Q–A metric, where \tool{}$_{MDPO}$ leads by a small margin of 0.01. Across all four metrics, the performance differences among these three approaches remain relatively small (average gap is smaller than 0.06), indicating that data quality plays a major role, while training objectives provide more fine-grained benefits.

Notably, \tool{} achieves the highest score on the S–A (Stylistic Strength) metric, validating the effectiveness of using style-specific rejected samples during DPO training. In contrast, \tool{}$_{MDPO}$ shows slightly inferior performance on S–A. A likely explanation is that its rejected samples differ significantly from the chosen ones across all four dimensions, allowing the model to easily distinguish them and optimize logits without focusing specifically on stylistic nuances. This diluted objective may have weakened the model’s ability to capture fine-grained stylistic preferences.

\subsection{Time Cost Analysis}
Among our baselines, SFT-prompt employs a smaller backbone model compared to R1-prompt, leading to shorter inference time. Therefore, we focus our runtime comparison on SFT-prompt and \tool{}. We measure the average inference time per sample on the test set: \tool{} takes 2.08 seconds, while SFT-prompt takes 2.47 seconds, resulting in a 1.19x speedup.
This improvement stems from the fact that in SFT-prompt, injecting style-related tokens significantly increases the input length, which imposes substantial overhead during decoding. In contrast, \tool{} injects stylistic knowledge via LoRA modules, which are lightweight and efficiently loaded. This comparison highlights the efficiency and scalability of parameterized style injection over prompt-based alternatives in latency-sensitive applications.
\subsection{Case Study}
\begin{figure}[t]
    \centering
    \includegraphics[width=\linewidth]{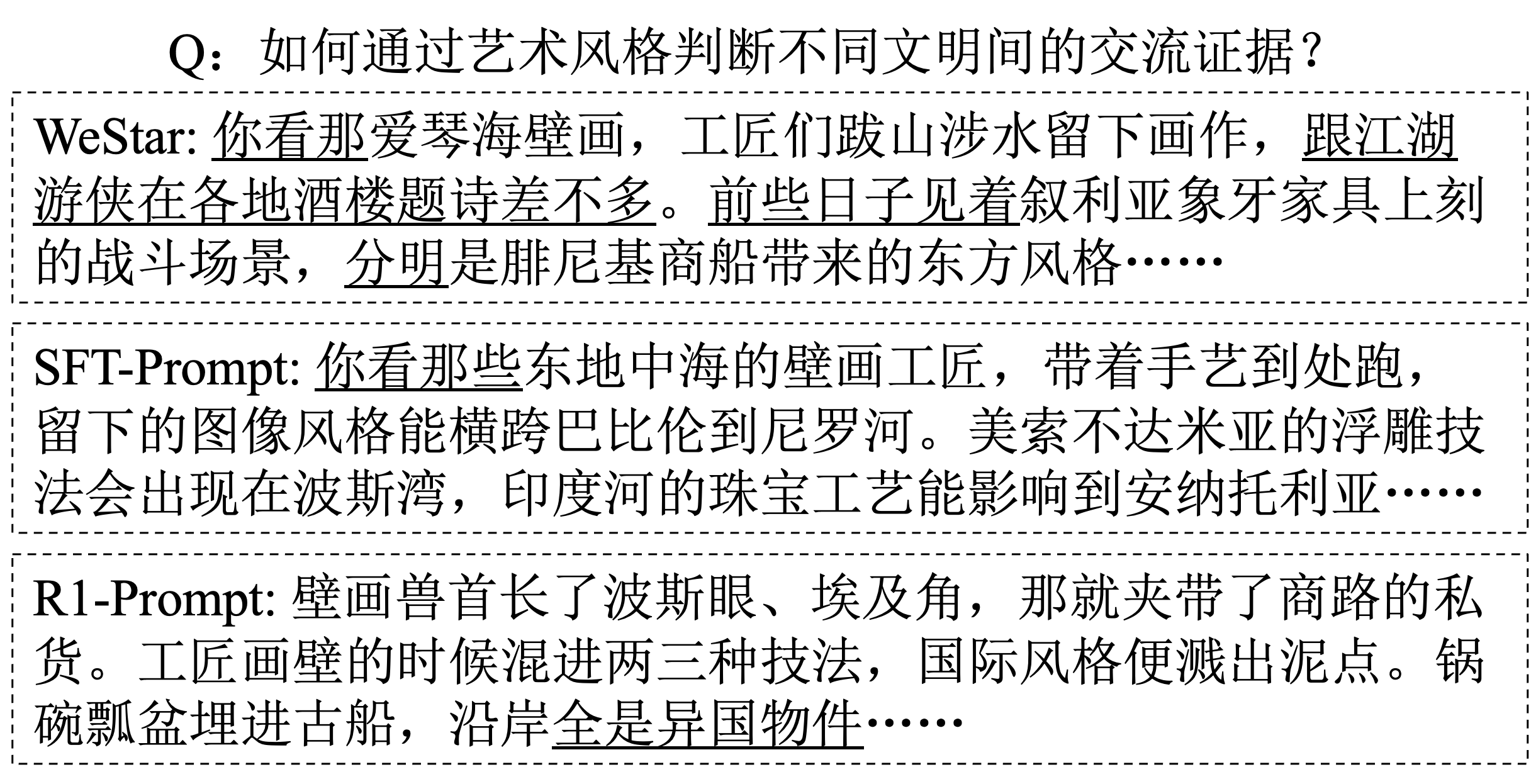}
    \vspace{-2mm}
    \caption{Cases of \tool{} vs. Prompt-Based Methods}
    \label{fig:case}
    \vspace{-4mm}
\end{figure}
We present a case study in Figure~\ref{fig:case}, comparing responses generated by \tool{} and two prompt-based methods for the same input question. The question is designed to reference three article snippets from an Official Account. To make the comparison intuitive and reader-friendly, we choose a stylistic exemplar from Chinese literature—the fictional character Huang Rong~\cite{huangrong} from Jin Yong's novels—as the target persona. Due to space limits, we omit the referenced article snippets and only display the opening portion of each generated response. Phrases, words, or sentence fragments that align with Huang's style are underlined. As shown, \tool{} produces responses that more consistently reflect the target style. Notably, the last sentence generated by \tool{} showed in Figure ~\ref{fig:case} omits the subject, which mirrors a grammatical-level stylistic trait of Huang’s speech patterns. In contrast, both prompt-based methods underperform in stylistic consistency. This is likely due to the injection of lengthy article segments into the prompt, which increases the context length and challenges the model’s attention window. This case highlights the effectiveness of \tool{}'s parameterized style representation: by encoding style-specific knowledge directly into the model's parameter space, \tool{} avoids the limitations imposed by prompt size and better preserves target style during generation.

\section{Conclusion}
In this work, we tackle the underexplored yet practical task of stylized contextual question answering for official accounts, where responses must be both style-aware and context-grounded at scale. 
Existing fine-tuning, CoT, and prompt-based methods struggle with efficiency or scalability in industrial settings. 
We propose \tool{}, a lite-adaptive AI assistant designed for millions of multi-style Official Accounts. 
\tool{} leverages RAG for retrieving context and PRAG for injecting style-specific LoRA modules. 
It builds high-quality training data through hierarchical style clustering, LLM-based stylized rewriting, and metric-driven selection, and further enhances stylistic alignment via Style-enhanced DPO. 
Experiments on large-scale industrial data show that \tool{} outperforms strong baselines in contextual relevance, style fidelity, and fluency—offering an effective, scalable solution for real-world deployments.

\bibliography{main}
\clearpage
\appendix
\end{document}